\pdfoutput=1
\documentclass[11pt]{article}
\usepackage[preprint]{acl}
\usepackage{times}
\usepackage{latexsym}
\usepackage[T1]{fontenc}
\usepackage[utf8]{inputenc}
\usepackage{microtype}
\usepackage{inconsolata}
\usepackage{graphicx}
\usepackage{siunitx,booktabs}
\usepackage{multirow}
\usepackage{tabularx}
\usepackage{relsize}
\usepackage{makecell}
\usepackage{stfloats}

\title{Improving Spoken Language Modeling with Phoneme Classification: \texorpdfstring{\\}{}A Simple Fine-tuning Approach}

\author{
  \textbf{Maxime Poli \textsuperscript{1}},
  \textbf{Emmanuel Chemla \textsuperscript{1}},
  \textbf{Emmanuel Dupoux \textsuperscript{1, 2}}
\\
  \textsuperscript{1}ENS - PSL, EHESS, CNRS
  \textsuperscript{2}Meta FAIR
\\
  \texttt{maxime.poli@ens.psl.eu}  
}

\begin{document}
\maketitle
\begin{abstract}
Recent progress in Spoken Language Modeling has shown that learning language directly from speech is feasible. Generating speech through a pipeline that operates at the text level typically loses nuances, intonations, and non-verbal vocalizations. Modeling directly from speech opens up the path to more natural and expressive systems. On the other hand, speech-only systems require up to three orders of magnitude more data to catch up to their text-based counterparts in terms of their semantic abilities. We show that fine-tuning speech representation models on phoneme classification leads to more context-invariant representations, and language models trained on these units achieve comparable lexical comprehension to ones trained on hundred times more data.
\end{abstract}

\section{Introduction and related work}
Recent advances in Self-supervised Speech Representation Learning (SSL) \cite{mohamed2022selfsupervised,chen2022wavlm,hsu2021hubert,baevski2020wav2vec} have enabled the development of label-free representations that are valuable for various downstream tasks \cite{yang2021superb}. These representations can be discretized and treated as pseudo-text, allowing for the training of language models directly from raw audio \cite{lakhotia-etal-2021-generative}, which capture both prosody and linguistic content \cite{kharitonov-etal-2022-text}. Applications of these audio-based language models include dialogue modeling \cite{nguyen-etal-2023-generative}, emotion conversion \cite{polyak2021speech}, and direct speech-to-speech translation \cite{lee-etal-2022-textless}. They can be trained not only on discretized SSL representations but also on continuous word-size tokens \cite{algayres-etal-2023-generative} or on a combination of acoustic and semantic tokens \cite{borsos2023audiolm}. However, these models still lag behind their text-based counterparts in terms of capturing semantics when trained with similar data quantity \cite{nguyen2020zero}, with scaling laws up to three orders of magnitude slower  \cite{cuervo2024scalingpropertiesspeechlanguage}. 
Recent approaches tackled this issue by jointly training speech and text Language Models (LMs) \cite{nguyen2024spiritlm,maiti2023voxtlm,chou-etal-2023-toward} or by using existing LMs as a warm initialization \cite{hassid2023textually}.

\begin{figure}
    \centering
    \includegraphics[width=\linewidth]{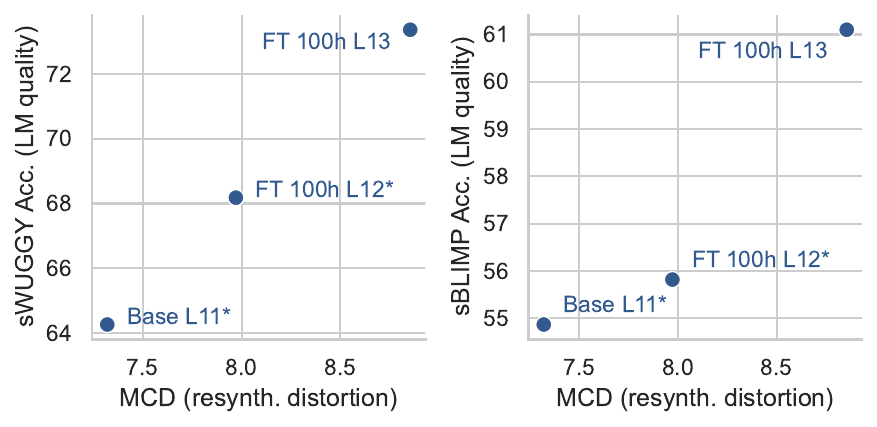}
    \caption{Trade-off between language modeling and expressive resynthesis. *: embeddings initialized from unit centroids.}
    \label{fig:tradeoff}
\end{figure}

\begin{figure*}
    \centering
    \includegraphics[width=\linewidth]{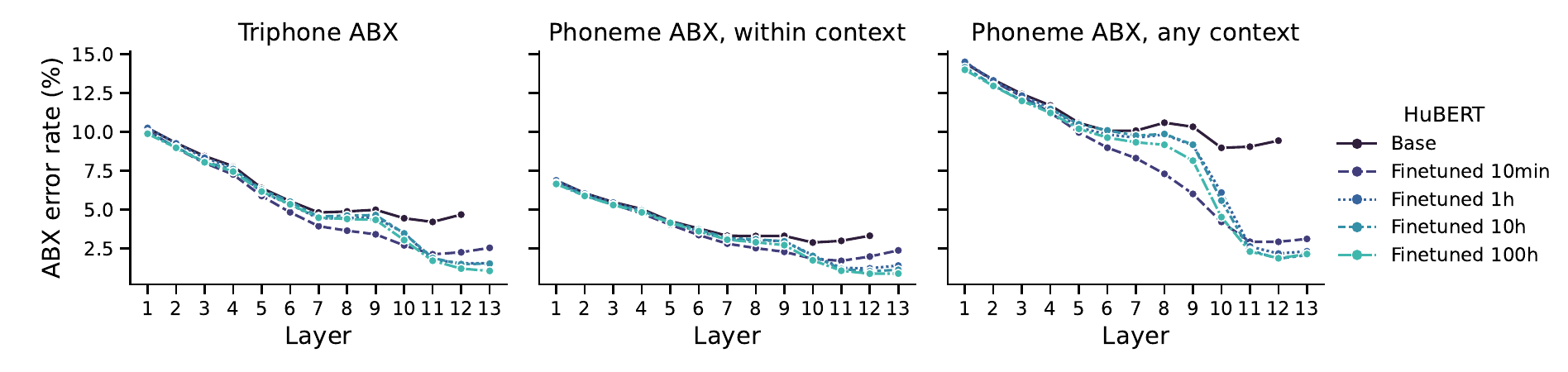}
    \caption{ABX error rate averaged across subset (dev-clean, dev-other) and speaker (within, across) conditions.}
    \label{fig:abxlayer}
\end{figure*}

One hypothesis for the data inefficiency of spoken language models is that they must at the same time perform language modeling and process irrelevant acoustic variations. Recent works have addressed this issue for background noise \cite{chen2022wavlm}, speech rate change \cite{gat-etal-2023-augmentation}, and speaker change \cite{qian2022contentvec,chang2023selfsupervised,chang-glass-2024-r}. However, contextual variations due to coarticulation remain a challenge \cite{hallap2023evaluating}: SSL units align more closely with contextual phone states \cite{young-etal-1994-tree} than with linguistic units \cite{dunbar2022selfsupervised}, which may affect the LM's capacity to learn higher-order representations of language.

Here, we test a simple idea: using supervised fine-tuning on a phoneme classification task to help the model remove its contextual dependency. We first show that fine-tuned models learn representations that are much more context-invariant than the original SSL representations, even with as little as a few hours of labels. Next, we show that these representations can be used to train a LM that outperforms the standard approach. We then evaluate whether the fine-tuned representations have retained their expressive power by measuring the distortion when resynthesizing expressive speech.

We release the code and models at \url{https://github.com/bootphon/spokenlm-phoneme}.

\section{Method}

\subsection{Phoneme classification}
We started from the pretrained HuBERT \cite{hsu2021hubert} Base model, with 95M parameters, and fine-tuned it on a frame-wise phoneme classification task with a forced aligned gold transcription. We chose this objective to give the model full information about phoneme identity and boundaries, to enforce the learning of context-invariant representations. An alternative would have been to use a CTC objective \cite{graves2006connectionist}, which has the advantage of not requiring forced-alignment, but may result in alignment errors hindering context-invariance. As shown in Appendix \ref{sec:ft}, CTC fine-tuning results in slightly lower performance than phone classification.

We added one fully connected layer on top of the HuBERT backbone that maps the 768-dimensional representation to our phoneme space of dimension 40. We fine-tuned this model on LibriSpeech \verb!train-clean-100! \cite{panayotov2015librispeech}. We also reported results for models fine-tuned on LibriLight Limited \SI{10}{\hour}, \SI{1}{\hour}, and \SI{10}{\minute} \cite{kahn2020librilight}. The forced alignments are those used in \citet{nguyen2020zero}, obtained with the Abkhazia library\footnote{\url{https://github.com/bootphon/abkhazia}}. The fine-tuning hyperparameters are derived from those used in \citet{hsu2021hubert} for ASR.
Input frames are partially masked as in pretraining, but the prediction loss is computed over all output frames, not just the masked ones. We trained for \num{20000} steps with a batch size of 32 on a single NVIDIA V100 GPU.

\subsection{Quantization}
We selected the best layer in terms of Triphone ABX score for the standard HuBERT base and the model fine-tuned on \verb!train-clean-100!. We trained k-means models on the features of a \SI{10}{\hour} subset of \verb|train-clean-100| extracted from those layers, with $k=500$. We also quantized the logits of the fine-tuned model by simply setting the labels as the predicted phonemes for each frame.

\subsection{Language modeling}
Finally, we trained LMs on the discretized units. The language model is a 3-layer LSTM, following the low-budget baseline of \citet{nguyen2020zero}, only changing the embedding dimension from $200$ to $768$. It was trained on the discrete units of LibriSpeech \SI{960}{\hour}, for \num{30000} steps on a single NVIDIA V100 GPU. This 26M parameters language model is two orders of magnitude smaller both in terms of number of parameters and hours of training data than Spoken LMs like TWIST \cite{hassid2023textually} or SpiRit-LM \cite{nguyen2024spiritlm}. Our fine-tuned units can in principle benefit any other LM, including these larger ones.

\subsection{Speech resynthesis}
For speech resynthesis, we trained a HiFi-GAN \cite{kong2020hifigan,polyak2021speech} on the \textsc{Expresso} dataset \cite{nguyen2023expresso}, conditioned on the HuBERT discrete speech units and one-hot speaker embeddings from one of \textsc{Expresso}'s voices. We trained for \num{250000} steps on two NVIDIA V100 GPUs and followed the other hyperparameters used in \textsc{Expresso}. In this setup the HiFi-GAN has a different training domain than the HuBERT, the k-means, and the LM, which were trained on the audiobooks of LibriSpeech. \textsc{Expresso} is rich in expressive variations, paralinguistics and nonvocals, making it well-suited to evaluate whether the discrete units preserve expressivity along with phonemic content.

\subsection{Evaluation metrics}
We evaluate continuous and discrete units using ABX discriminability \cite{schatz2013evaluating,schatz2016abxdiscriminability}. This task quantifies the discriminability between two sound categories, $A$ and $B$, as the probability that a token $x$ of category $A$ will be closer to another $a\in A$ than to a $b \in B$. The dissimilarity function is the dynamic time-warping aligned angular distance between the model's representations of two sounds. The ABX error rate is calculated by averaging the discriminabilities for all pairs of categories and subtracting it from $1$. In the standard evaluation, each token is a triphone and triphones differ only by the central phoneme in a triplet. In the ``within speaker'' condition, $a$, $b$, and $x$ come from the same speaker, while in the ``across speaker'' condition, $a$ and $b$ come from the same speaker, and $x$ from another one.

Following \citet{hallap2023evaluating}, we also evaluate our models on the Phoneme ABX task, where each token is a phoneme.  We examine two conditions: ``within context'' (constant preceding and following phonemes) and ``any context'' (no constraints on context). This task assesses context-invariance in speech representations, revealing that current self-supervised systems struggle with context independence. Notably, in \citet{hallap2023evaluating} the performance drop when removing the constant context condition is larger than the gaps observed in speaker independence or clean versus less-clean speech conditions. By fine-tuning at a frame level without taking into account the context, our approach is a way to directly tackle this issue. For complementary analysis of the discrete units, see Appendix \ref{sec:units}.

\begin{table}
    \centering
    \small
    \addtolength{\tabcolsep}{-1.5pt}
    \begin{tabular}{lccc}
\toprule
& \multirow{2.4}{*}{\makecell{Triphone \\ ABX $\downarrow$}} & \multicolumn{2}{c}{Phoneme ABX $\downarrow$} \\
\cmidrule(lr){3-4}
 & & W/in ctx & Any ctx  \\
\midrule
\qquad{\it Continuous} \\
wav2vec 2.0 Base L6 & 5.41 & 3.78 & 11.55 \\
WavLM Base L11 & 3.57 & 2.54 & 8.26 \\
ContentVec\textsubscript{100} L12 & 3.84 & 2.54 & 6.89\\
HuBERT + Spin\textsubscript{2048} L12 & 3.05 & 2.31 & 7.63 \\
\midrule
\qquad{\it Continuous} \\
Base L11 & 4.20 & 2.98 & 9.04 \\
FT 100h L12 & \underline{1.20} & \textbf{0.87} & \textbf{1.87} \\
FT 100h L13 & \textbf{1.05} & \underline{0.88} & \underline{2.14} \\
\qquad{\it Centroid} &  \\
Base L11 & 4.54 & 3.84 & 7.34  \\
FT 100h L12 & 1.65 & 1.92 & 2.76 \\
\qquad{\it One-hot} &  \\
Base L11 & 7.81 & 12.23 & 30.00  \\
FT 100h L12 & 4.02 & 6.51 & 26.88 \\
FT 100h L13 & 4.08 & 4.78 & 5.40 \\

\bottomrule
\end{tabular}
    \caption{ABX error rate on selected layers averaged across subset and speaker conditions. Without quantization, when considering the k-means centroid and with one-hot encoding. For each condition, the best score is in \textbf{bold} and the second best is \underline{underlined}.}
    \label{tab:abx}
\end{table}

\begin{table*}
    \centering
    \begin{tabular}{lcccccccc}
\toprule
    & \multicolumn{5}{c}{WER $\downarrow$} & & MCD $\downarrow$\\
    \cmidrule(lr){2-6} \cmidrule(lr){8-8}
      & dev-clean & dev-other & test-clean & test-other & \textsc{Expresso-read} && \textsc{Expresso} \\
\midrule
Original audio & 1.69 & 3.55 & 1.86 & 3.89 & 11.90 && - \\
Base L11 & 3.82 & 11.37 & 4.12 & 11.26 & 20.93 && 7.32 \\
FT 100h L12 & 4.36 & 10.75  & 4.62 & 10.90  & 23.03 && 7.97 \\
FT 100h L13 & 5.78 & 11.90 & 5.97 & 12.12 & 23.80 && 8.85 \\
\bottomrule
\end{tabular}

    \caption{Resynthesis evaluation. WER is computed using a wav2vec 2.0 ASR system on the resynthetized output. MCD compares the cepstral representation of the inputs and outputs.}
    \label{tab:resynthesis}
\end{table*}

\begin{table}
\captionsetup{belowskip=-2ex}
\centering
\addtolength{\tabcolsep}{-5pt}
\begin{tabular}{lccc}
\toprule
& \multicolumn{2}{c}{\textsc{sWUGGY} $\uparrow$} & \textsc{sBLIMP $\uparrow$} \\
\cmidrule(lr){2-3}
& all & in-vocab & \\
\midrule
GSLM (6k h) & - & 68.7 & 57.1 \\
AudioLM (60k h) & 71.5 & 83.7 & 64.7 \\
TWIST-7B (150k h) & 74.6 & 84.4 & 62.1 \\
\midrule
Base L11 (1k h)& 64.26 & 70.87 & 54.87 \\ 
FT 100h L12 (1k h) & 68.18 & 77.55 & 55.82 \\
FT 100h L13 (1k h) & 73.37 & 85.20 & 61.10 \\
\multicolumn{2}{l}{\quad \textit{Init from centroids}} \\
Base L11 (1k h)& 64.78 & 71.56 & 54.83 \\
FT 100h L12 (1k h) & 68.85 & 78.66 & 56.17 \\
\midrule
Gold phonemes (1k h) & 81.58 & 94.75 & 62.77 \\
\bottomrule
\end{tabular}
\caption{Zero-shot language comprehension scores (in \%), for LMs with an embedding table either initialized randomly or from the unit centroids.}
\label{tab:slm}
\end{table}

We evaluate spoken language modeling at the lexical and syntactic levels using the sWUGGY and sBLIMP metrics from the ZeroSpeech 2021 challenge \cite{nguyen2020zero}. sWUGGY is a ``spot-the-word'' task, where the network is presented with a word and a matching non-word, and evaluated on its ability to assign a higher probability to the true word. We also report results for ``in-vocab'' pairs, which only contains words from LibriSpeech. sBLIMP assesses the network's ability to prefer grammatically correct sentences over incorrect ones, given a pair of matching sentences.

We evaluate content preservation in resynthesized speech by following \cite{nguyen2023expresso} and running wav2vec 2.0 Large ASR \cite{baevski2020wav2vec} on the resynthesized speech, reporting the Word Error Rate (WER). We assess this on \textsc{Expresso-read} the reading subset of \textsc{Expresso} -- in-domain for the vocoder but out-of-domain for the HuBERT backbone and the k-means module -- and on LibriSpeech, which is out-of-domain for the vocoder. On \textsc{Expresso} the target voice is the same as the input voice, while on LibriSpeech the target voice is sampled from the four voices. We also compute the mel cepstral distortion
(MCD) \cite{kubichek1993mcd} between the original and resynthesized samples of \textsc{Expresso-read} using \citet{Sternkopf_mel-cepstral-distance_2024}.

\section{Results}

\subsection{Results at the phonemic level}
\label{sec:abx}

As shown in Figure \ref{fig:abxlayer}, we computed the ABX error rate for each Transformer layer of the base model and the fine-tuned models, including the added fully connected layer (layer 13). We calculated both triphone- and phoneme-level ABX error rates. Fine-tuning mainly improves the last layers' ABX error rates, with near-perfect scores for the 10h and 100h fine-tuned models in the ``within context'' condition. SSL representations generally struggle more in the ``any context'' condition: there the gain in error rate is the most significant, dropping from 9.4\% to 2.4\% after fine-tuning on as little as 10 minutes. Fine-tuning pushes representations to become more context-independent.

We selected the best layers for the base model (layer 11) and fine-tuned 100h model (layer 12) based on the Triphone ABX score, as well as the last layer of the fine-tuned 100h model (layer 13). We trained k-means on these representations and report the results in Table \ref{tab:abx}. We compare these to the ABX error rates of the best layers of wav2vec 2.0 \cite{baevski2020wav2vec}, WavLM \cite{chen2022wavlm}, ContentVec\textsubscript{100} \cite{qian2022contentvec} and HuBERT + Spin\textsubscript{2048} \cite{chang2023selfsupervised}. For the centroid scores, each representation is replaced by the continuous representation of the closest centroid in k-means. For the one-hot scores, each representation is replaced by a one-hot vector with a 1 at its label position. We use the same distance to compute the ABX as for continuous representations. In the case of the base model's layer 11 (Base L11) and the fine-tuned 100h model's layer 12 (FT 100h L12), the representations are of dimension 768, while for the fine-tuned 100h model's layer 13 (FT 100h L13) they have a dimension of only 40. Fine-tuning improves both triphone and phoneme ABX scores, particularly in reducing the context effect in the ``any context'' condition, as observed earlier.
In the case of the ABX of one-hot representations, the error rates increase across all conditions, but the highest increase is when the context is not shared between the phones in the triplet. This is a sign that the k-means clusters not only are organized according to the phonemes but also to the surrounding context. Clusters are grouped according to their most probable phoneme, and within each group, clusters encode different contexts. By going from centroid representations to one-hot representations, all 500 clusters are now equidistant, which leads to the dramatic loss in ``any context'' compared to the more modest ones in the other two conditions.

\subsection{Results above the phonemic level}

We report in Table \ref{tab:slm} the zero-shot sWUGGY (lexical level) and sBLIMP (syntactic level) scores for the base and fine-tuned models, as well as for an LSTM trained on the gold phonemes. Following the observation regarding the ABX error rates of the centroids, which remained within 1 percentage point of the standard continuous units, we train LSTMs by initializing their embedding table directly with the associated centroid representation of dimension 768. Apart from this change, the training process is the same between the two conditions.
Fine-tuning for phoneme classification improves spoken language modeling in terms of zero-shot comprehension evaluations. Overall, the gap between training from speech and training with golden phonemes is now halved. Fine-tuning for phoneme classification results in models that are on par in terms of lexical comprehension with much larger baselines, which were trained on orders of magnitude more of data. 

However, Table \ref{tab:resynthesis} shows that this comes at the cost of the quality of resynthesis. Notably, there is a cost in content preservation, illustrated by the WER. It exists both for the LibriSpeech dataset and for the \textsc{Expresso-read}, while these two datasets correspond to the training domain of different components of our pipeline. Figure \ref{fig:tradeoff} makes directly visible the trade-off between language modeling and speech generation quality.

\section{Conclusion}
We showed that fine-tuning SSL representations with a phoneme classification task is an effective and simple procedure to improve context independence. LMs trained on these units achieve comparable lexical comprehension to models trained on hundred times more data. And we also found that initializing the embeddings of the discrete tokens of the LMs with the centroids of the units further helps with LM scores. This shows that the units found are meaningfully placed relative to one another in this representation space. Our work also highlights the trade-off between language modeling (which works best with abstract units), and speech generation (which works best with specific units). Fine-tuning on phoneme classification can adjust this trade-off.

\section{Limitations}
Further work is needed to improve on the trade-off, perhaps by combining SSL, resynthesis, and fine-tuning objectives concurrently. More comprehensive studies could explore the role of the encoder in the spoken language modeling pipeline by examining the impact of fine-tuning methods on downstream language modeling, comparing self-supervised and supervised speech models with different kinds of supervision. Another important direction to consider is the application of this method in a multilingual setting. The benefits of fine-tuning are visible after training on as little as a few hours of aligned data, making it applicable to low resource languages.

\section*{Acknowledgments}
This work was performed using HPC resources from GENCI-IDRIS (Grant 2023-AD011014368) and was supported in part by the Agence Nationale pour la Recherche (ANR-17-EURE-0017 Frontcog, ANR10-IDEX-0001-02 PSL*, ANR19-P3IA-0001 PRAIRIE 3IA Institute) and a grant from CIFAR (Learning in Machines and Brains) awarded to E.D. in his EHESS capacity. M. P. acknowledges Ph.D. funding from Agence de l’Innovation de Défense.
\bibliography{anthology,custom}

\clearpage
\appendix

\begin{table*}[b]
    \centering
    \begin{tabular}{lccccc}
\toprule
      & dev-clean & dev-other & test-clean & test-other & \textsc{Expresso-read} \\
\midrule
Original audio  & 2.07 & 3.76 & 2.03 & 3.91 & 3.33 \\
Base L11 & 3.84 & 11.61 & 4.03 & 11.38 & 6.58  \\
FT 100h L12 & 4.24 & 10.97 & 4.34 & 10.67 & 7.95 \\
FT 100h L13 & 5.72 & 11.76 & 5.68 & 11.84 & 9.68 \\
\bottomrule
\end{tabular}
    \caption{WER using Whisper large-v3 (in \%)}
    \label{tab:wer_whisper}
\end{table*}

\begin{figure*}[ht]
    \centering \includegraphics[width=\linewidth]{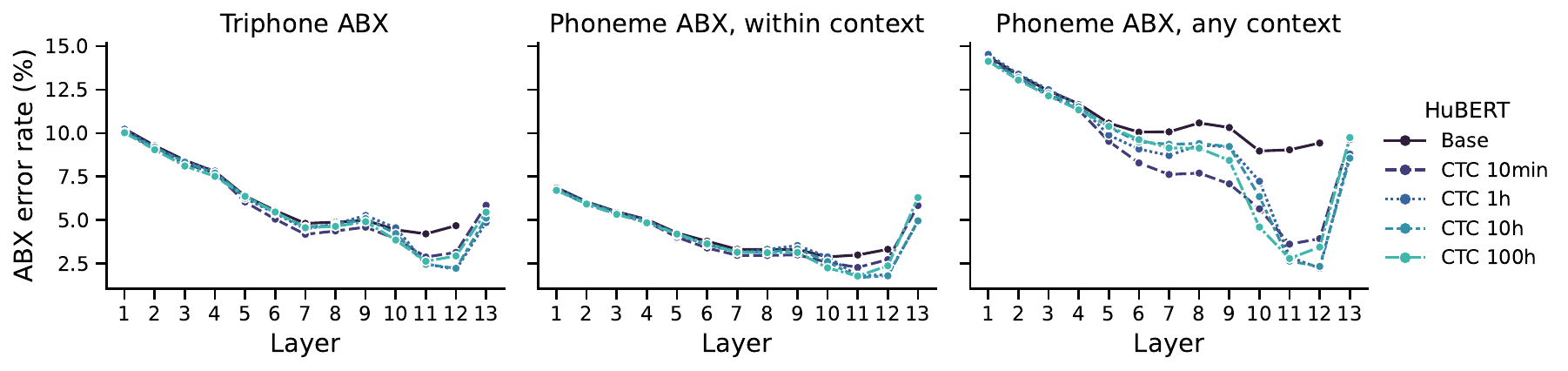}
    \caption{ABX error rate for models finetuned with CTC, averaged across subset and speaker conditions.}
    \label{fig:abxlayerctc}
\end{figure*}

\section{Appendix}
\subsection{Fine-tuning results}
\label{sec:ft}
As preliminary investigation, we fine-tuned HuBERT on phoneme recognition with a CTC loss instead of frame-level classification.  As shown in Figure \ref{fig:abxlayerctc} this results in slightly weaker performance in terms of ABX.

\begin{table}[ht]
\centering
\small
\addtolength{\tabcolsep}{-2.5pt}
\begin{tabular}{lcccc}
\toprule
 & dev-clean & dev-other & test-clean & test-other \\
\midrule
\multicolumn{5}{c}{\textit{Frame Classification Accuracy} $\uparrow$}\\
\textbf{Frame level} & & & & \\
10min & 88.80 & 83.78 & 88.80 & 84.29 \\
1h & 91.36 & 87.35  & 91.24 & 87.66  \\
10h & 93.01 & 89.03 & 92.96 & 89.31 \\
100h & 94.36 & 90.36 & 94.28 & 90.75 \\
\midrule
\multicolumn{5}{c}{\textit{Phone Error Rate} $\downarrow$}\\
\textbf{Frame level} & & & & \\
10min & 8.45 & 15.82 & 8.87 & 15.30 \\
1h & 4.68 & 9.59 & 5.15 & 9.25 \\
10h & 3.64 & 8.70& 4.02 & 8.38\\
100h & 2.83 & 7.53 & 3.15 & 7.07 \\
\\
\textbf{CTC} & & & & \\
10min & 8.27 & 15.18 & 8.73 & 14.72 \\
1h & 4.68 & 9.37 & 5.14 & 8.98 \\
10h & 3.27 & 7.29 & 3.65 & 7.00 \\
100h & 2.35 & 6.33 & 2.59 & 5.91 \\
\bottomrule
\end{tabular}
\caption{Fine-tuning results (in \%)}
\label{tab:accuracy_per}
\end{table}

Table \ref{tab:accuracy_per} presents the frame-level accuracy and Phone Error Rate (PER) for models fine-tuned on increasing labeled data quantity. The PER was computed by deduplicating consecutive predictions, without using a Language Model. For reference, the HuBERT base in SUPERB \cite{yang2021superb}, trained with the CTC objective and with a frozen backbone, has a PER of $5.41\%$ on \verb!test-clean!.

\subsection{Discrete units quality}
\label{sec:units}
In addition to the ABX scores reported in Section \ref{sec:abx}, the quality of the discrete units and their relationship to phonemes can also be assessed with the three metrics proposed in \citet{hsu2021hubert}: Cluster Purity, Phone Purity, and PNMI. Cluster purity is the conditional probability of a k-means label given a phone label, phone purity is the conditional probability of a phone label given a k-means label, and PNMI is the phone-normalized mutual information between units and phone labels. The units are obtained from the cluster assignments given by the k-means with 500 clusters trained on the output of the considered model. The evaluation is done on the combination of LibriSpeech \verb|dev-clean| and \verb|dev-other|. We have for the Base L11 and FT 100h L12 models: a PNMI of 0.669 and 0.846, Cluster Purity of 0.093 and 0.131, and Phone Purity of 0.685 and 0.858, respectively.

\subsection{Resynthesis evaluation with another ASR system}
We report Table \ref{tab:wer_whisper} the Word Error Rate for resynthesis on the evaluation datasets using Whisper large-v3 \cite{pmlr-v202-radford23a} instead of wav2vec 2.0 as the ASR system. The differences between models are consistent with those in Table \ref{tab:resynthesis}.

\clearpage

\subsection{Resynthesis quality by expressive style}

\begin{figure}
    \centering
    \includegraphics[width=\linewidth]{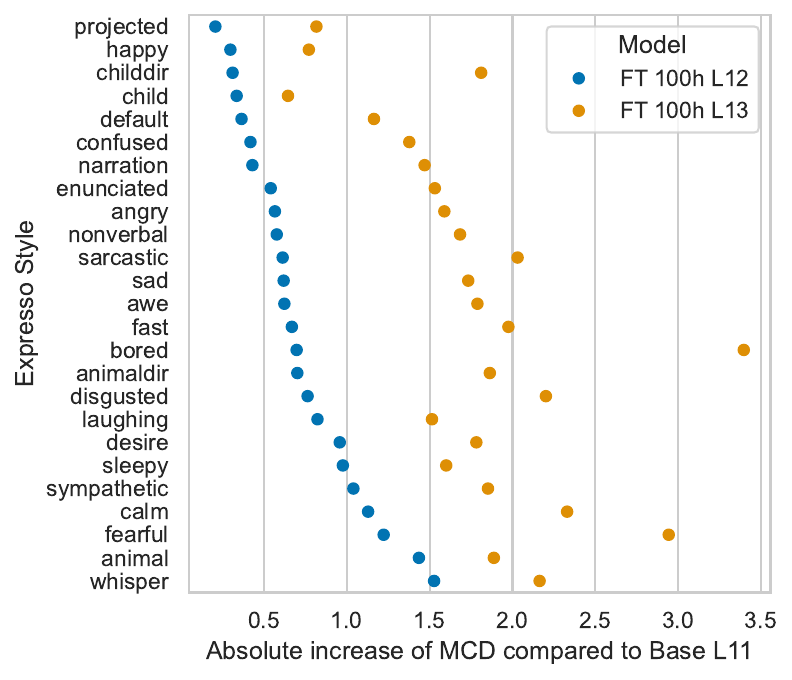}
    \caption{Difference between the MCD of the fine-tuned models and Base L11 on \textsc{Expresso} for each style.}
    \label{fig:mcdstyle}
\end{figure}

The drop in resynthesis quality by going from the standard model to the fine-tuned ones is further detailed is Figure \ref{fig:mcdstyle}. For each expressive style in \textsc{Expresso}, the fine-tuned models exhibit a higher MCD compared to Base L11. The difference is the most prominent for styles capturing more non-verbal vocalizations such as ``whisper'' or ``bored''.

\end{document}